\documentclass[3p,times,procedia]{elsarticle}
\usepackage{ecrc}

\usepackage{amsmath,amssymb,amsfonts}

\usepackage{graphicx}
\usepackage[figuresright]{rotating}
\usepackage{booktabs}
\usepackage{array}

\usepackage{xcolor}
\usepackage{textcomp}

\usepackage{tikz}
\usetikzlibrary{shapes, arrows, positioning ,calc}
\usepackage{pgfplots}
\pgfplotsset{compat=1.18}

\usepackage{caption}
\usepackage{subcaption}
\usepackage[numbers]{natbib}
\usepackage{hyperref}
\hypersetup{colorlinks=true, citecolor=blue}

\usepackage{listings}

\biboptions{authoryear}

\volume{00}
\firstpage{1}

\journalname{Transportation Research Procedia}
\runauth{A.A Adepitan et al}
\jid{trpro}

\hypersetup{colorlinks,
  linkcolor=blue,
  citecolor=blue,
  urlcolor=blue}

\begin{document}

\begin{frontmatter}

\dochead{The 2nd International Conference on Smart Mobility and Logistics Ecosystems (SMILE 2026)}%

\title{Learning Minimally-Congested Drive Times from Sparse Open Networks: A Lightweight RF-Based Estimator for Urban Roadway Operations.}

\author[1]{Adewumi Augustine Adepitan\corref{cor1}}
\ead{aadepita@gmu.edu}
\ead{adepitanadewumi@gmail.com}

\author[2]{Christopher J. Haruna}
\ead{christopher.haruna@bringvzw.be}

\author[3]{Morayo Ogunsina}
\ead{mogunsin@gmu.edu}

\author[4]{Damilola Olawoyin Yussuf.}
\ead{g202110610@kfupm.edu.sa}

\author[3]{Ayooluwatomiwa Ajiboye}
\ead{aajiboye@gmu.edu}

\address[1]{Sid and Reva Dewberry Department of Civil, Environmental, and Infrastructure Engineering, George Mason University, 4400 University Dr, Fairfax, VA 22030, USA}
\address[2]{Brussels Research and Innovation Center for Green Technologies (BRING VZW), Brussels, Belgium}
\address[3]{Department of Computer Science, George Mason University, Fairfax, VA 22030, USA}
\address[4]{Department: Electrical engineering, KFUPM}

\begin{abstract}
Accurate roadway travel-time prediction is foundational to transportation systems analysis, yet widespread reliance on either data-intensive congestion models or overly naïve heuristics limits scalability and practical adoption in engineering workflows. This paper develops a lightweight estimator for minimally-congested car travel times that integrates open road-network data, speed constraints, and sparse control/turn features within a random forest framework to correct bias from shortest-path traversal-time baselines. Using an urban testbed, the pipeline: (i) constructs drivable networks from volunteered geographic data; (ii) solves Dijkstra routes minimizing edge traversal time; (iii) derives sparse operational features (signals, stops, crossings, yield, roundabouts; left/right/slight/U-turn counts); and (iv) trains a regression ensemble on limited high-quality reference times to generalize predictions beyond the training set. Out-of-sample evaluation demonstrates marked improvements over traversal-time baselines across mean absolute error, mean absolute percentage error, mean squared error, relative bias, and explained variance, with no significant mean bias under minimally congested conditions and consistent k-fold stability indicating negligible overfitting. The resulting approach offers a practical middle ground for transportation engineering: it preserves point-to-point fidelity at metropolitan scale, reduces resource requirements, and supplies defensible performance estimates where congestion feeds are inaccessible or cost-prohibitive, supporting planning, accessibility, and network performance applications under low-traffic operating regimes.
\end{abstract}


\cortext[cor1]{Corresponding author. Tel.: 
1-202-699-3775 Adewumi Augustine Adepitan, George Mason University, Fairfax, VA 22030, USA. Email: aadepita@gmu.edu ; adepitanadewumi@gmail.com}

\end{frontmatter}

\section{Introduction}

Travel time prediction is fundamental to transportation engineering, supporting applications from accessibility analysis and network assessment to route planning and infrastructure investment. However, the field is divided between computationally intensive, data‑rich methods and oversimplified heuristics that fail to capture real‑world travel dynamics \cite{jenelius2013travel}.

State‑of‑the‑art models leverage massive GPS records, proprietary sensor feeds, or deep learning architectures, achieving MAPE values typically between 3\% and 17\% \cite{hou2018network}. Although accurate, these approaches demand substantial computational resources, technical expertise, and costly data, limiting their accessibility for many practitioners \cite{giles2022creating}. Conversely, planning‑oriented methods often rely on Euclidean distance, network distance, or speed‑limit‑based traversal times. These naïve approaches systematically underestimate actual travel times by ignoring traffic controls, turns, and intersection delays, leading to MAPE errors exceeding 20\% and potentially flawed planning decisions.

This paper bridges the gap by introducing a lightweight random‑forest estimator that uses sparse open data to predict minimally congested travel times with accuracy approaching state‑of‑the‑art methods while maintaining the accessibility and efficiency of simplified approaches. Our method builds on the premise that travel time depends on five key factors: distance, speed limits, traffic controls, turning movements, and congestion. While comprehensive congestion data remain proprietary, the first four components are widely available from open sources.

The core contribution is a systematic framework that transforms sparse open network data into accurate predictions through feature engineering and machine‑learning correction. By training on a limited set of high‑quality reference times and operational features derived from open street data, our approach achieves prediction quality comparable to resource‑intensive methods with modest computational requirements and freely available inputs.

Extensive validation in a major urban area demonstrates substantial improvements over baseline methods across multiple accuracy metrics. The resulting methodology offers a practical tool for generating metropolitan‑scale travel‑time estimates, especially where resource constraints or data limitations preclude the use of more sophisticated congestion‑aware systems.
\section{Related Work}

\subsection{Advanced Travel Time Prediction Methods}

Travel time prediction underpins transportation applications like accessibility analysis, route planning, and investment decisions. The field faces a methodological gap between data‑intensive advanced methods and oversimplified heuristics \cite{jenelius2013travel}. State‑of‑the‑art models leverage extensive GPS, sensors, or deep learning, achieving MAPE of 3–17\% \cite{hou2018network,vankdoth2023deep}, yet require substantial resources and proprietary data, limiting accessibility \cite{giles2022creating}. Conversely, planning‑oriented methods often rely on Euclidean or network distance, ignoring traffic controls and turns, leading to systematic underestimation and MAPE above 20\%. 

We bridge this gap with a lightweight random‑forest estimator that uses sparse open data to predict minimally congested travel times with accuracy near state‑of‑the‑art. Our approach rests on five travel‑time factors: distance, speed limits, traffic controls, turning movements, and congestion. The first four are openly available, while congestion data remain largely proprietary.

We present a systematic framework that transforms sparse open network data into accurate predictions through feature engineering and machine‑learning correction. Training on limited high‑quality reference times and open‑street features yields prediction quality comparable to costly methods with modest computational demands.

Validation in a major city demonstrates substantial improvements over baselines. The methodology offers a practical tool for metropolitan‑scale travel‑time estimation where resource or data constraints preclude sophisticated congestion‑aware systems.
\subsection{Simplified Methods in Planning and Practice}

Transport planning practice often prioritizes implementability over accuracy, employing simplified travel time estimation methods due to resource constraints. The simplest approaches minimize Euclidean distance, ignoring network structure and speeds. More sophisticated methods minimize network distance using sources like OpenStreetMap but neglect speed variations and operational delays.

A significant improvement incorporates speed limits to minimize traversal time, the sum of segment lengths divided by their speed limits. While capturing basic speed variations, this approach systematically underestimates actual travel times by ignoring intersection controls, turning movements, and acceleration/deceleration effects \cite{scott2008role}, typically underestimating urban travel by 15–25\% \cite{salonen2013modelling}.

Some researchers use traffic analysis zone (TAZ) matrices from planning organizations, offering empirical grounding but with coarse spatial resolution that obscures point‑to‑point variations \cite{levine2012does}. The gap between resource‑intensive methods and inadequately accurate simplified approaches presents a significant challenge. Our work addresses this by developing an intermediate approach that balances accuracy with practical implementability.
\subsection{Open Data in Transportation Modeling}

Due to increasing availability of open geographic data, such modeling can now be carried out without dependence on proprietary datasets. OpenStreetMap (OSM) has become an important resource for detailed data on the road network across the world.  Notwithstanding the quality and completeness of OSM data vary widely by region, especially for attributes like traffic controls and turn restrictions \cite{delmelle2019travel}.

Many researchers have applied statistical correction and feature engineering to improve open data.  Speed profiles based on OSM are developed by Ludwig et al. \cite{ludwig2023traffic} for improved travel time estimation in open routing services. In like manner, Yiannakoulias et al. \cite{yiannakoulias2013estimating} showed that the integration of turn penalties with basic signal information can accomplish better accessibility measurements.

Our approach extends these ideas by leveraging sparse open data in a systematic machine learning framework to refine traversal time estimates.  We achieve levels of accuracy suitable for transportation planning applications while eliminating the need for comprehensive real-time congestion data by concentrating on minimally congested situations.

\begin{figure}[h]
\centering
\begin{tikzpicture}[scale=0.9, transform shape, node distance=1.8cm] 
\tikzstyle{block} = [rectangle, draw, fill=blue!10,
    text width=3.5em, text centered, rounded corners, minimum height=2.5em, font=\footnotesize] 
\tikzstyle{line} = [draw, -latex', thick]

\node [block] (osm) {OSM\\Network};
\node [block, right of=osm] (graph) {Graph\\Build};
\node [block, right of=graph] (od) {OD\\Sampling};
\node [block, right of=od] (dijkstra) {Dijkstra\\Paths};
\node [block, right of=dijkstra] (features) {Feature\\Extract};
\node [block, right of=features] (rf) {RF\\Training};
\node [block, right of=rf] (prediction) {Predictions};

\node [block, below=0.8cm of dijkstra] (google) {Google\\API Ref.};

\path [line] (osm) -- (graph);
\path [line] (graph) -- (od);
\path [line] (od) -- (dijkstra);
\path [line] (dijkstra) -- (features);
\path [line] (features) -- (rf);
\path [line] (rf) -- (prediction);
\path [line] (google) -- (rf);

\end{tikzpicture}
\caption{Workflow integrating open network data, feature extraction, and reference times for random forest training.}
\label{fig:workflow}
\end{figure}
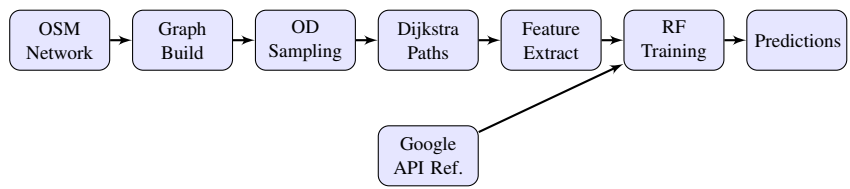
\section{Methodology}

\subsection{Conceptual Framework}

The starting point for our approach to predicting travel time is the observation that actual travel time \(t_{actual}\) incurs a systematic penalty with respect to a naive estimate \(t_{naive}\), and that this penalty is attributable to operational factors that are partially visible in open data. We visualize this relationship as.

\begin{equation}
t_{actual} = f(t_{naive}, \mathbf{X}_{controls}, \mathbf{X}_{turns}, \epsilon)
\end{equation}

In this formulation, \(\mathbf{X}_{controls}\) refers to the traffic control elements present along the corridor, \(\mathbf{X}_{turns}\) denotes the turning movement characteristics, and \(\epsilon\) signifies unobserved influences, including the random congestion effect.

The traversal time, denoted as \(t_{naive}\), is the estimate provided by Dijkstra’s algorithm. It was used to compute the path that minimizes the sum of traversal times on each street segment at speed limits. Though this baseline integrates the configuration of the network and fixed speed, it does not account for the delays progressively accumulated at intersections and the penalties incurred due to turning. This is a major fault as these factors severely impact actual travel time. More so in urban areas, which are peppered with intersections frequently, this would hold true \cite{yiannakoulias2013estimating}. 

Our approach rectifies this baseline via feature engineering and statistical learning. This is possible since many operational factors affecting travel time have open network data available, albeit infrequently and incompletely. The correction task appears suitable for the random forest algorithm \cite{breiman2001random}, which can accommodate the nonlinear relations, interactions and missing data typical of volunteered geographic information.
As depicted in Figure~\ref{fig:workflow} the pipeline of the proposed framework

\subsection{Data Acquisition and Processing}
We applied our methodology to the Los Angeles urban area for its complex network and available validation data. The study area was defined as the convex hull around Los Angeles County and urban area boundaries from the Global Human Settlement Layer's Urban Center Database, focusing on the core urbanized region.

Road network data were extracted from OpenStreetMap using OSMnx, keeping only the strongly connected component for routability. The network contained 63,359 tagged traffic control elements across five categories: crossings (21,560), stop signs (26,304), traffic signals (15,262), mini roundabouts (44), and give way signs (189). Sparse annotation coverage presents both a challenge and opportunity.

Origin‑destination pairs were generated via stratified sampling. From 5,000,000 random node pairs at intersections and dead‑ends, we filtered to realistic trips using Uber Movement hourly flow data. OD pairs matching real‑world trips at 3:00 AM (n=41,378) were retained to approximate minimally congested conditions.

Reference travel times were obtained from the Google Maps Routes API using the “BEST\_GUESS” model at 3:00 AM, when traffic is lowest and aligns with our free‑flow focus. The API supplied the fastest path, travel time, and length for each OD pair. Of 41,378 queries, 41,360 were resolved and used as the response variable.

\paragraph{Addressing route mismatch.} Features were extracted along the OSM Dijkstra shortest path. When Google's fastest route differed, the model learns to adjust the OSM path's time toward Google's reported time for a potentially different route. Both paths connect the same OD pair and share similar traffic controls and turn patterns, the features we capture, enabling the model to learn delay penalties despite slight sequence differences.

\subsection{Feature Engineering}

The feature set for travel time prediction comprised three categories: baseline traversal time, traffic control elements, and turning movements. Each category contributed distinct information for correcting the naïve travel time estimate.

The baseline feature (\(t_{naive}\)) was computed by applying Dijkstra's algorithm to find the path minimizing the sum of edge traversal times, where traversal time for each edge was calculated as length divided by speed limit. This provided the fundamental travel time estimate that would be obtained through conventional network analysis without operational corrections.

Traffic control features were derived by counting occurrences of five control types along each computed route: stop signs, traffic signals, pedestrian crossings, give way signs, and mini roundabouts. These counts captured the frequency of mandatory stops and delays along the route, with different control types expected to impose varying delay penalties.

The features of the turning movement quantified the angular deflection at each node along the route, categorized into five types according to established transportation engineering conventions \cite{salonen2013modelling}: left turns (90 ° -135 °), slight left turns (45° - 90 °), right turns (90 ° -135 °), slight right turns (45° - 90 °), and U-turns ($>$ 135 °). Straight movements (0°-45°) served as the reference category and were not explicitly included to avoid multicollinearity.

The complete feature matrix \(X\) thus contained 11 predictors: naïve travel time, five traffic control counts, and five turning movement counts. This feature set represented all operational factors derivable from open data that influence travel time under minimally-congested conditions.

\subsection{Model Specification and Training}
We formulated travel time prediction as a regression problem with the objective of learning a function \(f:\mathbb{R}^{11}\rightarrow\mathbb{R}\) that maps feature vectors to travel time predictions. The general model specification follows:
\[
t_{actual}=f(t_{naive},counts_{controls},counts_{turns})+\epsilon
\]

We evaluated four candidate algorithms for implementing \(f\): decision trees, random forests, gradient boosting, and AdaBoost. Model selection and hyperparameter tuning employed a standardized 80/20 training/test split with 5-fold cross-validation, optimizing for mean absolute error (MAE) through randomized grid search.

The random forest algorithm demonstrated superior performance and was selected as the final prediction model. Its ensemble approach, combining multiple decorrelated decision trees, proved particularly effective for handling the sparse and heterogeneous feature patterns derived from open data. The optimized hyperparameter configuration used 400 decision trees, a maximum depth of 10, bootstrap sampling with replacement, consideration of all features at each split, and a minimum of two samples required to split internal nodes.

We assessed model robustness through an additional 5-fold cross-validation on the complete dataset, with resulting MAE values of 75.3, 74.5, 73.6, 73.2, and 74.7 seconds, demonstrating consistent performance across folds and indicating negligible overfitting.

\paragraph{Reproducibility.} Our analysis used the Google Maps Routes API with parameters: departure time = 3:00 AM, traffic model = BEST\_GUESS, region = Los Angeles. 
Network characteristics and traffic control elements are summarized in Table~\ref{tab:performance}.

\section{Experimental Results}

\subsection{Model Performance Comparison}

We evaluated our random forest model against the naïve traversal time baseline and other machine learning approaches using six accuracy metrics. Table \ref{tab:performance} presents out‑of‑sample validation results for all candidate models.

The random forest model achieved a MAPE of 8.41\%, a substantial improvement over the naïve baseline MAPE of 21.15\%. This performance falls within the range of state‑of‑the‑art methods (3–17\% MAPE) while using only sparse open data and modest computational resources. Gradient boosting achieved a slightly lower MAPE (7.86\%) but exhibited significant under‑prediction bias.

\begin{figure}[h]
\centering
\begin{tikzpicture}[scale=0.85] 
\begin{axis}[
    width=2.8in, 
    height=2.0in, 
    ybar,
    bar width=12pt, 
    xlabel={Error Metric},
    ylabel={Value},
    symbolic x coords={MAPE, MAE, MSE, $\delta$},
    xtick=data,
    ymin=0,
    legend style={
        at={(0.5,-0.25)}, 
        anchor=north, 
        legend columns=-1,
        font=\footnotesize 
    },
    tick align=outside,
    tick pos=left,
    scaled ticks=false,
    label style={font=\small},
    tick label style={font=\footnotesize}
]
\addplot coordinates {(MAPE,21.15) (MAE,183.68) (MSE,48214.06) ($\delta$,182.85)};
\addplot coordinates {(MAPE,8.41) (MAE,75.32) (MSE,12154.79) ($\delta$,0.38)};
\legend{Naïve Baseline, RF Model}
\end{axis}
\end{tikzpicture}
\caption{Comparison of error metrics between naïve baseline and random forest model, showing substantial improvement.}
\label{fig:errors}
\end{figure}
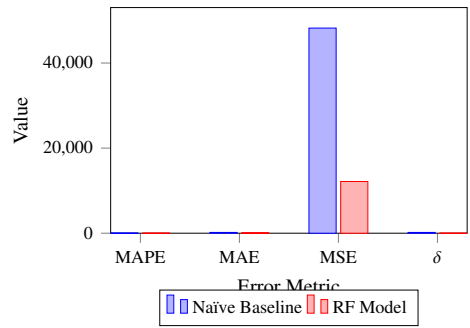

In absolute terms, the random forest reduced MAE from 183.68 to 75.32 seconds an improvement factor of 2.4 and MSE from 48,214.06 to 12,154.79 square seconds, a 4.0‑fold reduction. Figure \ref{fig:errors} visualizes these improvements.

Difference‑in‑means \(t\)-tests confirmed the naïve baseline’s systematic under‑prediction (\(\delta = -182.85\)s, \(p<0.01\)), while the random forest showed no significant mean bias (\(\delta = 0.38\)s, \(p=0.76\)). The average pairwise ratio (APR) improved from 0.79 (21\% under‑prediction) to 1.01 (1\% over‑prediction), and \(R^2\) increased from 0.74 to 0.93, indicating substantially greater explained variance.
\subsection{Algorithm Comparison and Selection}

Comparative analysis of the four candidate algorithms revealed important performance trade-offs. While all machine learning approaches substantially outperformed the naïve baseline, they exhibited distinct characteristics in terms of bias-variance properties.

Gradient boosting achieved the lowest MAPE (7.86\%) and MAE (71.99 seconds) but demonstrated significant systematic under-prediction (\(\delta = -19.20\) seconds, \(p < 0.01\)). This bias suggests potential over-regularization or insufficient complexity to fully capture the correction required from the baseline estimate. Similarly, AdaBoost showed competitive error metrics but significant under-prediction (\(\delta = -9.76\) seconds, \(p < 0.01\)).

The decision tree model exhibited minimal mean bias (\(\delta = 0.13\) seconds, \(p = 0.92\)) but higher variance, as evidenced by its superior MSE (13,570.17 square seconds) relative to the ensemble methods. This performance pattern aligns with theoretical expectations regarding the high variance of individual decision trees.

The random forest approach achieved an optimal balance between bias and variance, with minimal systematic prediction bias and competitive error metrics across all measures. Its ensemble structure effectively mitigated the high variance of individual trees while maintaining flexibility to capture complex feature relationships. Based on this comprehensive performance profile, we selected random forest as our final prediction model.

\subsection{Feature Importance Analysis}

To get the contribution of different feature categories to prediction of travel time, we conducted feature importance analysis based on the inbuilt importance metrics of random forest model. The findings indicated that naive travel time was the main predictor, with an R-squared of around 68\%. The baseline estimate is useful, as this result shows. However, improving the estimation with operational features can lead to much better results.

Among traffic control elements, traffic signals emerged as the most influential predictor, followed by stop signs and crossings. The minimal importance of roundabouts and give way signs likely reflects their sparse distribution within the network rather than negligible operational impact. Turning movement features collectively contributed approximately 19\% of the model's explanatory power, with left and right turns showing similar importance levels and U-turns demonstrating minimal influence due to their rarity in computed routes.

The feature importance pattern aligns with transportation engineering principles, confirming that intersection controls and turning movements constitute significant sources of delay beyond basic traversal time. Importantly, even sparse and incomplete operational data provided sufficient signal for meaningful prediction improvement, supporting the feasibility of our open data approach.

\begin{table}[h]
\centering
\caption{Prediction accuracy of candidate models}
\label{tab:performance}
\begin{tabular}{lcccc}
\toprule
Model & MAPE (\%) & MAE (s) & $\delta$ (s) & $R^2$ \\
\midrule
Naïve TT      & 21.2 & 184 & -182.9 & 0.74 \\
RF (Ours)     & \textbf{8.41} & \textbf{75} & \textbf{0.38} & \textbf{0.93} \\
GB            & 7.86 & 72 & -19.2 & 0.93 \\
DT            & 9.00 & 80 & 0.13 & 0.93 \\
AdaBoost      & 8.20 & 74 & -9.76 & 0.93 \\
\bottomrule
\end{tabular}
\end{table}

\section{Discussion}

\subsection{Practical Implications for Transportation Engineering}

The development of accurate, accessible travel time prediction methods has significant implications for transportation engineering practice. Our approach addresses a critical gap in the methodological landscape by providing a solution that balances accuracy requirements with practical implementation constraints. The 8.41\% MAPE achieved by our model represents a substantial improvement over commonly used simplified methods while avoiding the resource-intensive requirements of state-of-the-art approaches.

For transportation planning applications, accurate travel time estimates are essential for accessibility analysis, environmental justice assessments, and infrastructure investment prioritization. Systematic under-prediction of travel times, as exhibited by naïve methods, can lead to overstated accessibility measures and potentially inequitable resource allocation \cite{tsou2005accessibility}. Our method's minimal systematic bias ensures that planning decisions reflect actual travel conditions more accurately, supporting more equitable and effective transportation investments.

The computational efficiency of our approach enables metropolitan-scale analyses on standard computing hardware, addressing a key limitation of many advanced methods. The ability to generate accurate point-to-point travel time matrices for large urban regions supports detailed spatial analysis that is often precluded by computational constraints or API cost limitations. This scalability makes our method particularly valuable for regional transportation planning and policy evaluation.

Furthermore, our reliance on open data sources enhances methodological transparency and reproducibility. Unlike black-box commercial APIs or proprietary data sources, the open components of our approach can be scrutinized, validated, and adapted to local contexts. This transparency supports evidence-based decision-making and facilitates methodological advancement through community collaboration.

\subsection{Limitations and Boundary Conditions}

Our approach has several limitations. The focus on minimally congested conditions restricts applicability to contexts where congestion effects are minimal; during peak periods, unmodeled congestion may cause travel times to exceed our predictions. However, for planning applications such as infrastructure capacity analysis, emergency response planning, and off‑peak accessibility assessment, minimally congested times remain valuable references.

Open data quality and completeness present another potential limitation. While OpenStreetMap provides broad network coverage, traffic‑control tagging varies regionally. Sparse tagging may degrade model performance due to reduced feature information, though our results demonstrate robustness to such sparsity.

The generalizability of our model parameters across different urban contexts requires further validation. Although the conceptual framework is universally applicable, the relationship between operational features and travel‑time delays may vary due to differences in driver behavior, enforcement practices, and intersection design. Adaptation to new regions may require additional reference data for calibration.

Finally, our approach does not explicitly model temporal variations beyond minimally congested conditions. Incorporating time‑dependent patterns would demand additional data sources and modeling complexity, potentially compromising the accessibility advantages central to our method.

\subsection{Future Research Directions}

There are many ways through which we can expand our approach. The first step is to employ the approach in the application of different temporal conditions by collecting reference data over long time frames. More detailed congestion modelling would require a lot more resources. But there is some possibility with slight extra reference data, coarse temporal differentiation (e.g. peak versus off-peak).

Another promising avenue is to integrate further open data. New types of data like crowd-sourced information on traffic conditions, public transport schedules and weather data may be used to enhance predictive accuracy without adopting a closed data model. In like manner, when the availability of traffic control data is limited, it may be possible to address this limitation through imputation methods in a spatial statistics or machine learning framework.

More regional model calibration procedures would make the methodology more generalizable while maintaining the methodology. Techniques in transfer learning could adapt models using little local reference data from data-rich zones to application in new zones with minor data collection.

In the end, real-time routing applications integration is a fascinating application domain. It may be possible to have routing systems that are accurate yet lightweight and very efficient with the computational resources to allow large-scale deployments by combining our accurate baseline predictions with our lightweight congestion proxies.

\section{Conclusion}

The present paper proposes a lightweight framework based on random forests so as to forecast minimally-congested travel times using sparse open data. Combining the network traversal time with operational features obtained from open street data achieves performance close to those of complex and resource-consuming methods, while remaining at least as efficient as simple methods.

Validation in a big city showed performance improvements over conventional methods of measuring the time between two peaks. MAPE improved from 21.15\% to 8.41\% and systematic prediction bias was entirely removed. Due to its capability to counter-balance bias and variance and its ability in dealing with the sparse and heterogeneous feature patterns of open geographic data, this random forest algorithm works fine.

Our methodology fills a gap in transportation engineering practice that contains neither a simple heuristic solution nor an advanced method computationally demanding solution. This middle-ground solution allows for accurate metropolitan-scale travel time analysis without any reliance on proprietary data or specialized computational resources, thus encouraging broader adoption of evidence-based transportation planning.

The methods and data sources we use are open, which ensures transparency and reproducibility, while the modular structure allows for adaptation and augmentation in many urban contexts. The improved quality of open data and the ease of access to computation resources means approaches like ours have the potential to enable advanced transportation analysis capabilities beyond organizations with a lot of resources.

Subsequent efforts must work on achieving a longer period coverage, using other databases, and designing transfer learning techniques to regionalize models. As we move forward, we will continue to bridge the gap between the sophistication of our methodologies and the implementability of these methodologies in practice. In doing so, we can evolve the science and practice of transportation engineering. In addition, this must be done for the benefit of the community that we serve.

\bibliographystyle{plain}  
\bibliography{bib}
\end{document}